%% file: main.tex
\documentclass[10pt,twocolumn,letterpaper]{article}

\usepackage{3dv}
\usepackage{times}
\usepackage{epsfig}
\usepackage{graphicx}
\usepackage{amsmath}
\usepackage{amssymb}
\usepackage{multirow}
\usepackage{booktabs}
\usepackage{caption}
\usepackage[pagebackref=true,breaklinks=true,letterpaper=true,colorlinks,bookmarks=false]{hyperref}

\newcommand{\figref}[1]{Fig.\ref{#1}}
\newcommand{\tabref}[1]{Tab.\ref{#1}}
\newcommand{\equref}[1]{Eqn.(\ref{#1})}
\newcommand{\secref}[1]{Sec.\ref{#1}}

\threedvfinalcopy 

\def\threedvPaperID{20} 

\ifthreedvfinal\fi
\begin{document}

\title{Dynamic Multi-Person Mesh Recovery From Uncalibrated Multi-View Cameras}

\author{Buzhen Huang\hspace{2mm}\hspace{5mm} Yuan Shu\hspace{2mm}\hspace{5mm} Tianshu Zhang\hspace{2mm}\hspace{5mm} Yangang Wang\footnotemark[1]\\%
\\
Southeast University, China
}

\twocolumn[{%
\renewcommand\twocolumn[1][]{#1}%
\maketitle
\thispagestyle{empty}
\begin{center}
   \centering
   \includegraphics[width=1.0\textwidth]{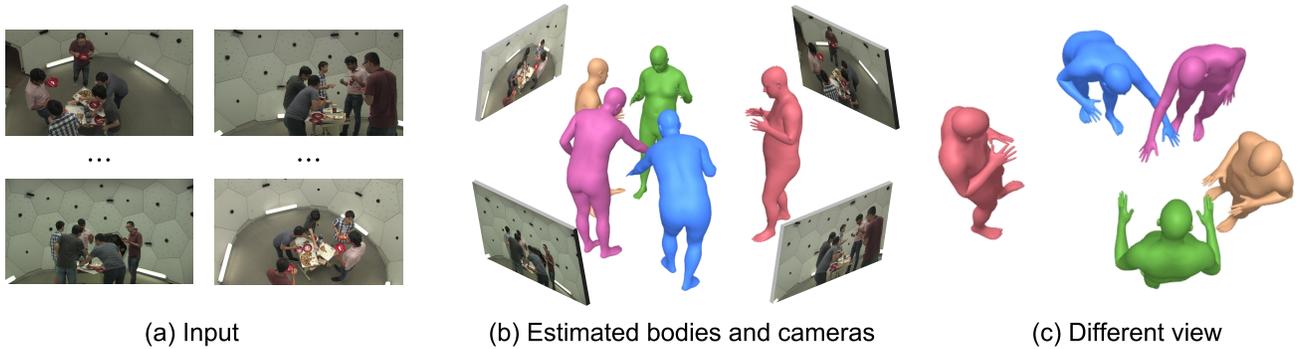}
   \vspace{-7mm}
   \captionof{figure}{Given multi-person video sequences from sparse uncalibrated cameras, our method simultaneously recovers human motions and extrinsic camera parameters from noisy human semantics.}
\end{center}%
}]

\begin{abstract}
   Dynamic multi-person mesh recovery has been a hot topic in 3D vision recently. However, few works focus on the multi-person motion capture from uncalibrated cameras, which mainly faces two challenges: the one is that inter-person interactions and occlusions introduce inherent ambiguities for both camera calibration and motion capture; The other is that a lack of dense correspondences can be used to constrain sparse camera geometries in a dynamic multi-person scene. Our key idea is incorporating motion prior knowledge into simultaneous optimization of extrinsic camera parameters and human meshes from noisy human semantics. First, we introduce a physics-geometry consistency to reduce the low and high frequency noises of the detected human semantics. Then a novel latent motion prior is proposed to simultaneously optimize extrinsic camera parameters and coherent human motions from slightly noisy inputs. Experimental results show that accurate camera parameters and human motions can be obtained through one-stage optimization. The codes will be publicly available at~\url{https://www.yangangwang.com}.
\end{abstract}

\renewcommand{\thefootnote}{\fnsymbol{footnote}}
\footnotetext[1]{Corresponding author. E-mail: yangangwang@seu.edu.cn. All the authors from Southeast University are affiliated with the Key Laboratory of Measurement and Control of Complex Systems of Engineering, Ministry of Education, Nanjing, China.}

\input{introduction.tex}

\input{relatedwork.tex}

\input{method.tex}

\input{experiments.tex}

\input{conclusion.tex}

{\small
\bibliographystyle{ieee}
\bibliography{egbib}
}

\end{document}

%% file: introduction.tex
\section{Introduction}
Recovering multiple human motions from video is essential for many applications, such as social behavior understanding, sports broadcasting, virtual reality applications, \etc. Numerous previous works have been aimed at capturing multi-person motions from multi-view input via geometry constraints~\cite{belagiannis20143d, dong2019fast, chen2020cross, lin2021multi, zhang20204d, joo2017panoptic} or optimization-based model fitting~\cite{ye2012performance, li2020full, liu2013markerless, li2018shape, wu2013set}. While these works have made remarkable advances in multi-person motion capture, they all rely on accurate calibrated cameras to build view-view and model-view consistency. Few works focus on multi-person motion capture from uncalibrated cameras. \cite{mustafa2015general} constructs a two-stage framework that first calibrates the camera using the static geometry from the background and then generates 3D human models from dynamic object reconstruction and segmentations. \cite{ershadi2021uncalibrated} utilizes the similarity of the estimated 3D poses in each view to find pose pairs and refines them in the global coordinate system. However, these methods require a large space distance among the target people and can not capture interactive human bodies.

In this paper, we address the problem of directly recovering multiple human bodies with unknown extrinsic camera parameters. There are two main challenges. The first one is that inter-person interactions and occlusions introduce inherent ambiguities for both camera calibration and motion reconstruction. The ambiguous low-level visual features lead to severe low and high frequency noises in detected human semantics~(\eg, 2D pose~\cite{belagiannis20153d}, appearance~\cite{li2020full}), which causes extreme difficulty in establishing view-view and model-view consistency. The other is that a lack of sufficient local image features~(\eg, SIFT~\cite{lowe2004distinctive}) can be used to constrain sparse camera geometries in a dynamic multi-person scene.

\begin{figure*}
    \begin{center}
    \includegraphics[width=0.95\linewidth]{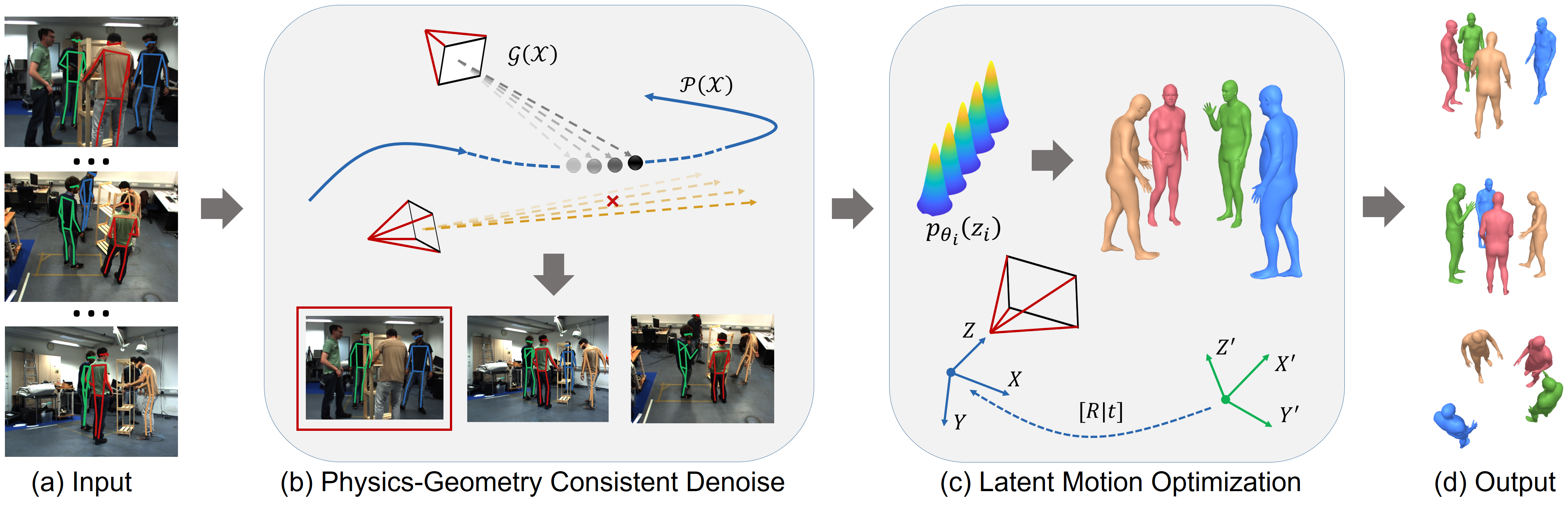}
    \end{center}
    \vspace{-8mm}
    \caption{Overview of our method. Since directly optimizing cameras and human motions from noisy detections~(a) always lead to suboptimal solutions, we first introduce a physics-geometry consistency~(b) to reduce high and low frequency noises in the detected human semantics. Then, to recover from the filtered partial and slightly noisy inputs~(b), we incorporate a novel latent motion prior to the optimization framework~(c) to obtain accurate camera parameters and coherent human motions~(d).}
    \label{fig:pipeline}
    \vspace{-6mm}
    \end{figure*}

To tackle the obstacles, our key-idea is to \textbf{use motion prior knowledge to assist the simultaneous recovery of camera parameters and dynamic human meshes from noisy human semantics.} We introduce a physics-geometry consistency to reduce the low and high-frequency noises of the detected multi-person semantics. Then a latent motion prior is proposed to recover multiple human motions with extrinsic camera parameters from partial and slightly noisy multi-person 2D poses. As shown in \figref{fig:pipeline}, the multi-view 2D poses from off-the-shelf 2D pose detection~\cite{fang2017rmpe,cao2019openpose} and tracking~\cite{zhou2019omni} contain high-frequency 2D joint jitter and low-frequency identity error. Without proper camera parameters, we can not filter out the noises by epipolar constraint~\cite{belagiannis20143d, chen2020cross}. However, we found that the triangulated skeleton joint trajectories are continuous, even though the camera parameters are inaccurate. Based on this observation, we propose a physics-geometry consistency and construct a convex optimization to combine kinetic energy prior and epipolar constraint to reduce the high and low frequency noises.

Simultaneously optimizing extrinsic camera parameters and multi-person motions from the filtered and slightly noisy 2D poses is a highly non-convex problem. We then introduce a compact latent motion prior to jointly recover temporal coherent human motions and accurate camera parameters. We adopt a variational autoencoder~\cite{kingma2013auto}~(VAE) architecture for our motion prior. Different from existing VAE-based motion models~\cite{lohit2021recovering, luo20203d, ling2020character}, we use bidirectional GRU~\cite{cho2014learning} as backbone and design a latent space both considering local kinematics and global dynamics. Therefore, our latent prior can be trained on a limited amount of short motion clips~\cite{mahmood2019amass} and be used to optimize long sequences. While the motion prior can generate diverse and temporal coherent motions, it is not robust to noises in motion optimization. We found that linearly interpolating the latent code of VPoser~\cite{pavlakos2019expressive} will produce consecutive poses. Inspired by this, we propose a local linear constraint on motion latent code in model training and optimization. This constraint ensures motion prior to produce coherent motions from noisy input. In addition, to keep local kinematics, a skip-connection between explicit human motion and latent motion code is incorporated in the model. Using the noisy 2D poses as constraints, we can recover human motions and camera parameters by simultaneously optimizing the latent code and cameras.

The main contributions of this work are summarized as follows.
\begin{itemize}
    \vspace{-3mm}
    \item We propose a framework that directly recovers multi-person human motions with accurate extrinsic camera parameters from sparse multi-view cameras.
   \vspace{-3mm}
    \item We propose a physics-geometry consistency to reduce the notorious low and high frequency noises in detected human semantics.
    \vspace{-3mm}
    \item We propose a human motion prior that contains both local kinematics and global dynamics, which can be trained on limited short motion clips and be used to optimize temporal coherent long sequences.
    \vspace{-3mm}
\end{itemize}

%% file: relatedwork.tex
\section{Related Work}\label{sec:relatedwork}
\noindent\textbf{Multi-view Human pose and shape estimation}. Reconstructing human pose and shape from multi-view inputs has been a long-standing problem in 3D vision. \cite{liu2013markerless} reconstructs interactive multi-person with manually specified masks. To avoid manual operations, the color \cite{mitchelson2003simultaneous,wu2013set}, appearance \cite{li2020full}, location \cite{li2018shape} and other cues of human are utilized to build the spatio-temporal correspondences, thus realizing optimization-based model fitting. In contrast, \cite{belagiannis20143d,belagiannis20153d,lin2021multi,zhang20204d,bridgeman2019multi,joo2017panoptic} firstly establish view-view correspondences via detected 2D poses and geometric constraints and then reconstruct through triangulation or optimization. \cite{dong2019fast} considers geometric and appearance constraints simultaneously. However, these methods all rely on accurate camera parameters. Besides, 2D poses and appearance can be easily affected by partial occlusion, which is very common in multi-person interaction sceneries.

To recover multiple human meshes from uncalibrated cameras, \cite{mustafa2015general} first calibrates the camera using the static geometry from the background and then generates 3D human models from dynamic object reconstruction. \cite{ershadi2021uncalibrated} realizes reconstruction via the similarity of the detected 3D poses from different views. However, these methods require a large space distance among the target people and can not capture interactive human bodies.

\noindent\textbf{Extrinsic camera calibration}. Conventional camera calibration methods rely on specific tools (\eg, checkerboard\cite{zhang2000flexible} and one-dimensional objects\cite{zhang2004camera}). Except for the complex calibration process, it leads to two separate stages for calibration and reconstruction. \cite{inomata2011registration,mustafa2015general,zou2012coslam} propose more convenient methods that directly use image features from static background (\eg, SIFT~\cite{lowe2004distinctive}) to calibrate the camera. However, the dynamic human bodies occupy the most proportion of the image pixels in multi-person scenarios. To handle this obstacle, \cite{sha2020end,cioppa2021camera,chen2019sports,sha2020end, citraro2020real} obtain structure cues and estimate camera parameters from the semantics of the scene~(\eg, lines of the basketball court). \cite{huang2016camera,tang2019esther} estimate the extrinsic camera parameters from the tracked human trajectories in more general multi-person scenes. \cite{sinha2010camera,ben2016camera,boyer2006using} extract frontier points of the silhouette and recover epipolar geometry by using points between different perspectives. Nevertheless, getting accurate human segmentations from in-the-wild images itself is a challenging problem. \cite{desai2018skeleton} realizes camera calibration by using the depth camera in an indoor scene to extract the skeleton. \cite{puwein2014joint,garau2019unsupervised,takahashi2018human} and \cite{garau2020fast} use detected human 2D joints and mesh respectively to calibrate the camera, further simplifying the calibration device. State-of-the-art 2D/3D pose estimation frameworks \cite{fang2017rmpe,cao2019openpose,kolotouros2019learning} can hardly get accurate 2D/3D keypoints in multi-person scenes, and such methods cannot be directly applied to multi-person cases. To reduce the ambiguities generated by human interactions and occlusions, we propose a physics-geometry consistent denoising framework and a robust latent motion prior to remove the noises, realizing multi-person reconstruction and extrinsic camera calibration in an end-to-end way.

\noindent\textbf{Motion prior}. Traditional marker-less motion capture relies on massive views to provide sufficient visual cues~\cite{joo2017panoptic, vlasic2009dynamic, collet2015high}. To reconstruct from sparse cameras, \cite{zhou2018monocap, li2020full} employ the euclidean distance of poses in adjacent frames as the regularization term, which may limit the dynamics of the reconstructed motions. Thus, applying strong and compact motion prior in motion capture has attracted wide attention. The simple and feasible motion priors~(\eg, Principal Component Analysis~\cite{sidenbladh2000stochastic}, Low-dimensional Non-linear Manifolds~\cite{jaeggli2009learning, gall20102d}) lack expressiveness and are not robust to noises. Historically, Gaussian Process Latent Variable Model~(GPLVM)~\cite{lawrence2005probabilistic, yao2011learning, li20103d, li2007simultaneous} succeed in modeling human motions~\cite{wang2007gaussian, urtasun20063d} since it takes uncertainties into account, but is difficult to make a smooth transition among mixture models. \cite{huang2017towards} uses low-dimensional Discrete Cosine Transform (DCT) basis~\cite{akhter2012bilinear} as the temporal prior to capture human motions. With the development of deep learning, VIBE \cite{kocabas2020vibe} trains a discriminator to determine the quality of motion, but one-dimensional variables can hardly describe dynamics. \cite{lohit2021recovering} and \cite{luo20203d, zhao2021travelnet} train VAEs based on Temporal Convolutional Networks(TCN) and Recurrent Neural Network(RNN) respectively and represent motion with latent code. However, both of these two methods use latent code in a fixed dimension, which is not suitable for dealing with sequences of varying lengths. \cite{ling2020character} constructs a conditional variational autoencoder~(cVAE) to represent motions of the two adjacent frames. Although this structure solves the problem of sequence length variation, it can only model sequence information of the past, which is not suitable for optimizing the whole sequence.

In this paper, we propose a motion prior that contains local kinematics and global dynamics of the motion. The structure of the model makes it is suitable for large-scale variable-length sequence optimization.

%% file: method.tex
\section{Method}\label{sec:method}
Our goal is to recover both multi-person motions and extrinsic camera parameters simultaneously from multi-view videos. Firstly, we propose a physics-geometry consistency to reduce the high and low frequency noises in the detected human semantics~(\secref{sec:denoising}). Then, we introduce a robust latent motion prior~(\secref{sec:motion prior}), which contains human dynamics and kinematics, to assist estimation from noisy inputs. Finally, with the trained motion prior, we design an optimization framework to recover accurate extrinsic camera parameters and human motions from multi-view uncalibrated videos~(\secref{sec:optimization}). 

\subsection{Preliminaries}\label{sec:Preliminaries}
\noindent\textbf{Human motion representation}.
We adopt SMPL~\cite{loper2015smpl} to represent human motion, which consists of the shape $\beta \in \mathbb{R}^{10}$, pose $\theta  \in \mathbb{R}^{72}$ and translation $\mathcal{T} \in \mathbb{R}^{3}$. To generally learn human dynamics and kinematics from training data, we separate global rotation $\mathcal{R} \in \mathbb{R}^{T \times 3}$, translation $\mathcal{T}$ and human shape $\beta$ when constructing the motion prior. Moreover, we use the more appropriate continuous 6D rotation representation~\cite{zhou2019continuity} for the prior. Finally, a motion that contains $T$ frames is represented as $\mathcal{X} \in \mathbb{R}^{T \times 138}$.

\noindent\textbf{2D pose detection and camera initialization}.
We first use off-the-shelf 2D pose estimation~\cite{fang2017rmpe} and tracking framework~\cite{zhou2019omni} to get tracked 2D poses for each person. Then, we estimate initial camera extrinsic parameters for the denoising framework~\secref{sec:denoising}. We obtain the fundamental matrix from multi-view 2D poses in the first frame using epipolar geometry with known intrinsic parameters. Then the initial extrinsic parameters can be decomposed from it. Since the 2D poses are noisy, a result selection is used to ensure robustness. The details can be found in the Sup. Mat.

\begin{figure}
    \begin{center}
    \includegraphics[width=1\linewidth]{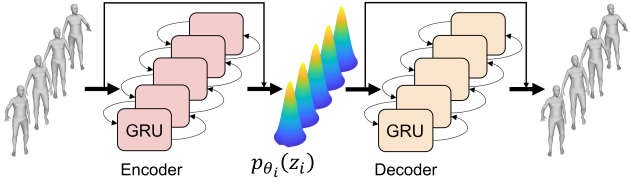}
    \end{center}
    \vspace{-6mm}
    \caption{The motion prior is a symmetrical encoder-decoder network, which compactly models human dynamics and kinematics. The prior can be trained on short clips and be used to fit long sequences.}
    \label{fig:motionprior}
    \vspace{-6mm}
    \end{figure}

\subsection{Physics-geometry Consistent Denoising}\label{sec:denoising}
Due to the inherent ambiguities in inter-person interactions and occlusions, state-of-the-art pose detection and tracking methods~\cite{fang2017rmpe, cao2019openpose, sun2019deep, zhou2019omni} can hardly get the precise 2D poses with accurate identity from in-the-wild videos. The drift and jitter generated by pose detection are often high-frequency, while identity error generated by pose tracking is low-frequency. The mixture of the two types of noises is notorious in multi-person mesh recovery. To solve this obstacle, we propose a physics-geometry consistency to reduce both high and low frequency noises in 2D poses from each view. 

Supposing the target person is detected in $V$ views, our goal is to remove the noisy detections that do not satisfy the physics-geometry consistency. Theoretically, despite that the camera parameters are not accurate, the triangulated skeleton joint trajectories from 2D poses with accurate identity are continuous. So we first utilize a set of optical rays, which come from the optical center of the camera and pass through corresponding 2D joint coordinates, to construct a physical constraint. For view $i$, the ray in the pl\"ucker coordinates is represented as $(n_{i}, l_{i})$. Given the skeleton joint positions of the previous frame $x_{t-1}$, the optical rays should be close to $x_{t-1}$. We represent the distance between $x_{t-1}$ and the rays as:

\begin{equation}
  \mathcal{L}^{i}_{p} = || x_{t-1} \times n_{i} - l_{i} ||.
\end{equation}

The rays generated by the wrong detection will produce an out-of-range physical cost $\mathcal{L}_{p}$. However, with only the above physical constraint, the system may get the wrong results in inter-person occlusion cases. Consequently, we further propose an additional geometric constraint. We enforce the rays from view $i$ and view $j$ to be coplanar precisely:

\begin{equation}
    \mathcal{L}_{g}^{i,j} = n_{i}^T l_{j} + n_{j}^T l_{i}.
\end{equation}

We combine these two constraints as the physics-geometry consistency. We then follow \cite{huang2013consistent} to filter out incorrect detections with the physics-geometry consistency. The physical cost and geometric cost of different views are represented in matrices $\mathcal{P}$ and $\mathcal{G}$.

\begin{equation}
    \left\{
        \begin{aligned}
        \mathcal{P}_{i,j} &= \mathcal{L}_{p}^{i} + \mathcal{L}_{p}^{j} \\
        \mathcal{G}_{i,j} &= \mathcal{L}_{g}^{i,j}
        \end{aligned}
    \right.,
\end{equation}
where $\mathcal{P}_{i,j}$ and $\mathcal{G}_{i,j}$ are physical cost and geometric cost of view $i$ and view $j$. We use a positive semidefinite matrix $\mathcal{M} \in \{0,1\}^{v \times v} $ to represent the correctness of correspondences among different views. Our goal is to solve $\mathcal{M}$, which minimizes the physics-geometry consistency cost:
\begin{equation}
    \mathop{\arg\min}_{\mathcal{M}} f(\mathcal{M})=-c_{g}\langle\mathcal{G}, \mathcal{M}\rangle-c_{p}\langle\mathcal{P}, \mathcal{M}\rangle,
\end{equation}
where $c_{g}$, $c_{p}$ are 0.7 and 0.3 in our experiment. $\langle \rangle$ denotes the hadamard product. Finally, we use the estimated $\mathcal{M}$ to extract accurate detections. 

The skeleton joint position of the start frame $x_0$ is triangulated with the queries of pose tracking~\cite{zhou2019omni}. We triangulate $x_t$ with filtered results and use it to calculate the physical consistency cost in the next frame. The filtered 2D poses will be used in \equref{equ:data} to find optimal motions. More details can be found in Sup. Mat.

\subsection{Latent Motion Prior}\label{sec:motion prior}
Simultaneous optimization of multi-person motions and camera parameters from slightly noisy 2D poses is a highly non-convex problem and is likely to fall into the local minima. To address this challenge, we design a compact VAE-based latent motion prior to obtain accurate and temporal coherent motions. The prior has three strengths. 1) It contains compact dynamics and kinematics to reduce computational complexity. 2) It can be trained on short motion clips and applied to long sequence fitting. 3) The latent local linear constraint ensures robustness to noisy input. The details are described as following.

\noindent\textbf{Model architecture}. 
Our network is based on VAE~\cite{kingma2013auto}, which shows great power in modeling motions~\cite{ling2020character, luo20203d}. As shown in~\figref{fig:motionprior}, the encoder consists of a bidirectional GRU, a mean and variance encoding network with a skip-connection. The decoder has a symmetric network structure. Different from previous work~\cite{ling2020character}, the bidirectional GRU ensures that the prior is able to see all the information from the entire sequence and that the latent code can represent global dynamics. However, the latent prior encoded only by features extracted from GRU is difficult to reconstruct accurate local fine-grained poses when used for large-scale sequence optimization. Thus, we construct a skip-connection for the encoder and decoder, respectively, allowing the latent prior to accurately capture the refined kinematic poses and the global correlation between them. Besides, we design the latent code $\mathbf{z} \in \mathbb{R}^{T \times 32}$ whose frame length $T$ is corresponding to the input sequence. Thus, our prior can be trained on a limited amount of short motion clips~\cite{mahmood2019amass} and be applied to long sequence fitting.

\noindent\textbf{Training}. 
In the training phase, a motion $\mathcal{X}$ is fed into the encoder to generate mean $\mu\left(\mathcal{X}\right)$ and variance $\sigma\left(\mathcal{X}\right)$. The sampled latent code $\mathbf{z} \sim q_{\phi}\left(\mathbf{z} \mid \mu\left(\mathcal{X}\right), \sigma\left(\mathcal{X}\right)\right)$ is then decoded to get the reconstructed motion $\mathbf{\hat{\mathcal{X}}}$. The reparameterization trick~\cite{kingma2013auto} is adopted to achieve gradient backpropagation. We train the network through maximizing the Evidence Lower Bound (ELBO):

\begin{equation}
    \begin{array}{l}
    \log p_{\theta}\left(\mathcal{X}\right) \geq \mathbb{E}_{q_{\phi}}\left[\log p_{\theta}\left(\mathcal{X}\mid\mathbf{z}\right)\right] \\
    -D_{\mathrm{KL}}\left(q_{\phi}\left(\mathbf{z}\mid\mathcal{X}\right) \| p_{\theta}\left(\mathbf{z}\right)\right).
    \end{array}
\end{equation}

The specific loss function is:
\begin{equation}
    \mathcal{L}_{vae} = \mathcal{L}_{\mathrm{6d}} + \mathcal{L}_{\mathrm{v}} + \mathcal{L}_{\mathrm{kl}}+ \mathcal{L}_{\mathrm{linear}}+\mathcal{L}_{\mathrm{reg}},
\end{equation}
where $\mathcal{L}_{6d}$ and $\mathcal{L}_{v}$ are:

\begin{equation}
    \mathcal{L}_{6d}=\sum_{t=1}^{T}\left\|\mathcal{X}_{t}-\mathbf{\hat{\mathcal{X}}}_{t}\right\|^{2},
\end{equation}

\begin{equation}
    \mathcal{L}_{v}=\sum_{t=1}^{T}\left\|\mathbf{\mathcal{V}}_{t}-\mathbf{\hat{\mathcal{V}}}_{t}\right\|^{2},
\end{equation}

where $\mathcal{V}_{t}$ is the deformed SMPL vertices of frame $t$. This term guarantees that the prior learns high fidelity local details.
\begin{equation}
    \mathcal{L}_{kl}=K L(q(\mathbf{z}\mid\mathcal{X}) \| \mathcal{N}(0, I)),
\end{equation}

which enforces its output to be near the Gaussian distribution. The regularization term, which ensures the network will not be easily overfitted:

\begin{equation}
    \mathcal{L}_{r e g}=\|\phi\|_{2}^{2}.
\end{equation}

Although applying the above constraints can produce diverse and temporal coherent motions, it is not robust to noisy 2D poses. The jitter and drift of 2D poses and identity error will result in an unsmooth motion. Inspired by the interpolation of VPoser~\cite{pavlakos2019expressive}, we add a local linear constraint to enforce a smooth transition on latent code:
\begin{equation}
    \mathcal{L}_{linear}=z_{t+1}-2z_{t}+z_{t-1}.\label{equ:linear}
\end{equation}

When the motion prior is applied in long sequence fitting, the parameters of the decoder are fixed. The latent code is decoded to get the motion $\hat{\mathcal{X}} \in \mathbb{R}^{T \times 138}$.

\subsection{Joint Optimization of Motions and Cameras}\label{sec:optimization}
 \noindent\textbf{Optimization variables}. Different from traditional structure-from-motion~(SFM), which lacks structural constraints between 3D points and is not robust to noisy input. We directly optimize the motion prior, so that the entire motions are under inherent kinematic and dynamic constraints. The optimization variables of $V$ views videos that contain $N$ people are $\{\left(\beta, \mathbf{z}, \mathcal{R}, \mathcal{T}\right)_{1: N}, \mathcal{E}_{1:V}\}$. The $\mathcal{E} \in \mathbb{R}^{6}$ is camera extrinsic parameter that contains rotation and translation.

\noindent\textbf{Objective}.
We formulate the objective function as following:
\begin{equation}
    \mathop{\arg\min}_{\left(\beta, \mathbf{z}, \mathcal{R}, \mathcal{T}\right)_{1: N}, \mathcal{E}_{1:V}} \mathcal{L}=\mathcal{L}_{data}+\mathcal{L}_{prior}+\mathcal{L}_{pen},
\end{equation}

where the data term is:
\begin{equation}
    \mathcal{L}_{data} =\sum_{v=1}^{V} \sum_{n=1}^{N} \sigma_{v}^{n} \rho\left(\Pi_{\mathcal{E}_{v}}\left(\mathrm{J}^{n}\right)-\mathbf{p}_{v}^{n}\right)\label{equ:data}
\end{equation}

where $\rho$ is the robust Geman-McClure function~\cite{geman1987statistical}. $\mathbf{p}$, $\sigma$ are the filtered 2D poses and its corresponding confidence. $\mathrm{J}$ is the skeleton joint position generated by model parameters.

Besides, the regularization term is:
\begin{equation}
    \mathcal{L}_{prior} = \sum_{n=1}^{N} \left\|\mathbf{z}_{n}\right\|^{2} + \sum_{n=1}^{N} \left\|\beta_{n}\right\|^{2} + \sum_{n=1}^{N} \mathcal{L}_{linear}.
\end{equation}

$\mathcal{L}_{linear}$ is the same as~\equref{equ:linear}. We further apply a collision term based on differentiable Signed Distance Field~(SDF)~\cite{jiang2020coherent} to prevent artifacts generated from multi-person interactions.

\begin{equation}
    \mathcal{L}_{pen}=\sum_{j=1}^{N} \sum_{i=1, i \neq j}^{N} \sum_{vt \in \mathcal{V}_{j}} -\min (\operatorname{SDF}_{i}(vt), 0),
\end{equation}
where $\operatorname{SDF}(vt)$ is the distance from sampled vertex $vt$ to the human mesh surface.

\begin{figure*}
    \begin{center}
    \includegraphics[width=0.95\linewidth]{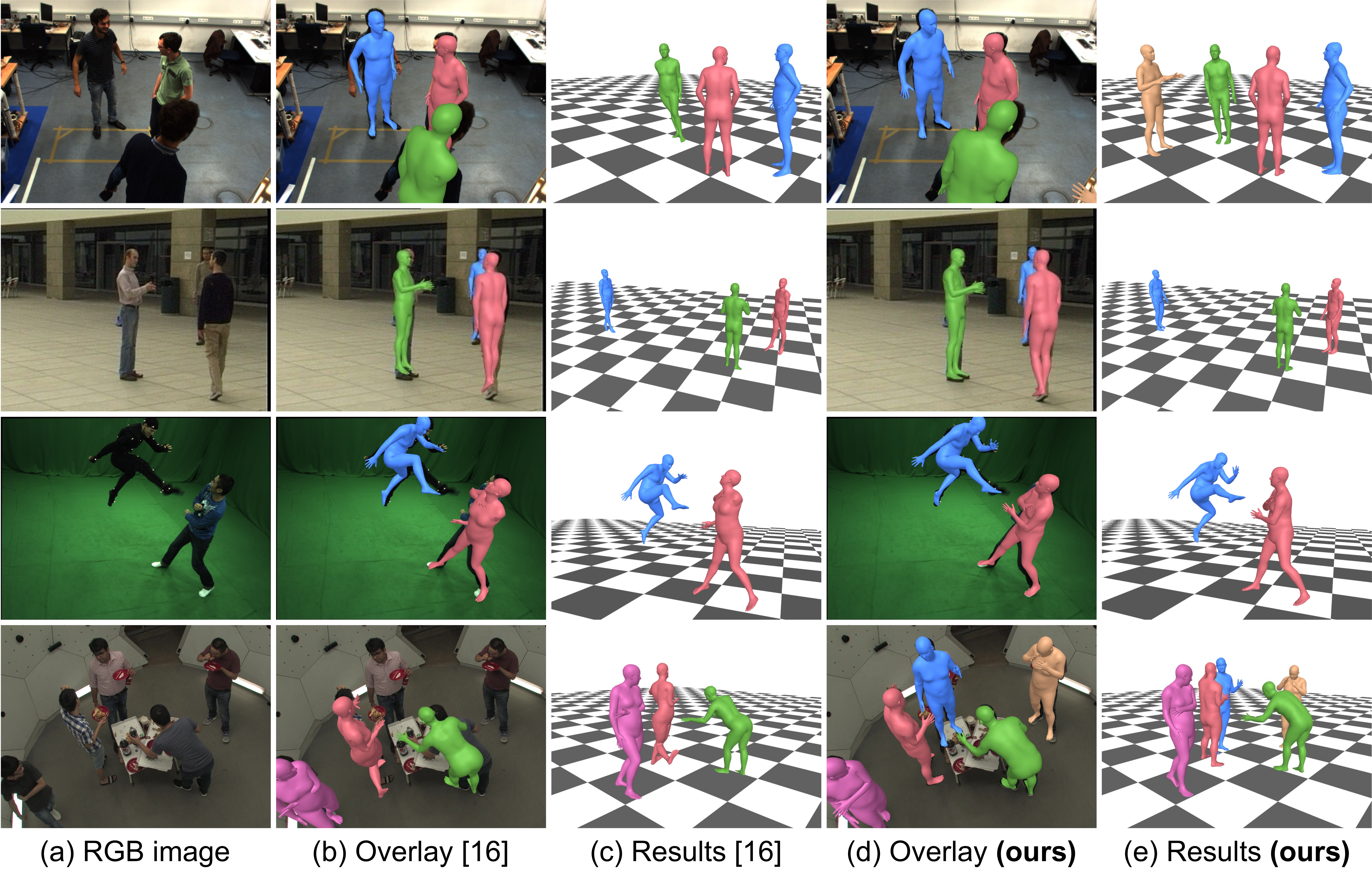}
    \end{center}
    \vspace{-7mm}
    \caption{Qualitative comparison with \cite{dong2019fast}. Due to the mismatched 2D pose and a lack of prior knowledge, \cite{dong2019fast} fails on these cases while our method obtains accurate results with the proposed motion prior and physics-geometry consistency.}
    \label{fig:qulitative_comparison}
    \vspace{-3mm}
    \end{figure*}

%% file: experiments.tex
\section{Experiments}\label{sec:experiment}
In this section, we conduct several evaluations to demonstrate the effectiveness of our method. The comparisons in \secref{sec:mocap experiment} show that our method can recover multiple human bodies from uncalibrated cameras and achieves state-of-the-art. Then, we prove that the accurate extrinsic camera parameters can be obtained from joint optimization. Finally, several ablations in \secref{sec:ablation} are conducted to evaluate key components. The details of the datasets that are used for training and testing can be found in the Sup. Mat.

\begin{figure*}
    \begin{center}
    \includegraphics[width=0.95\linewidth]{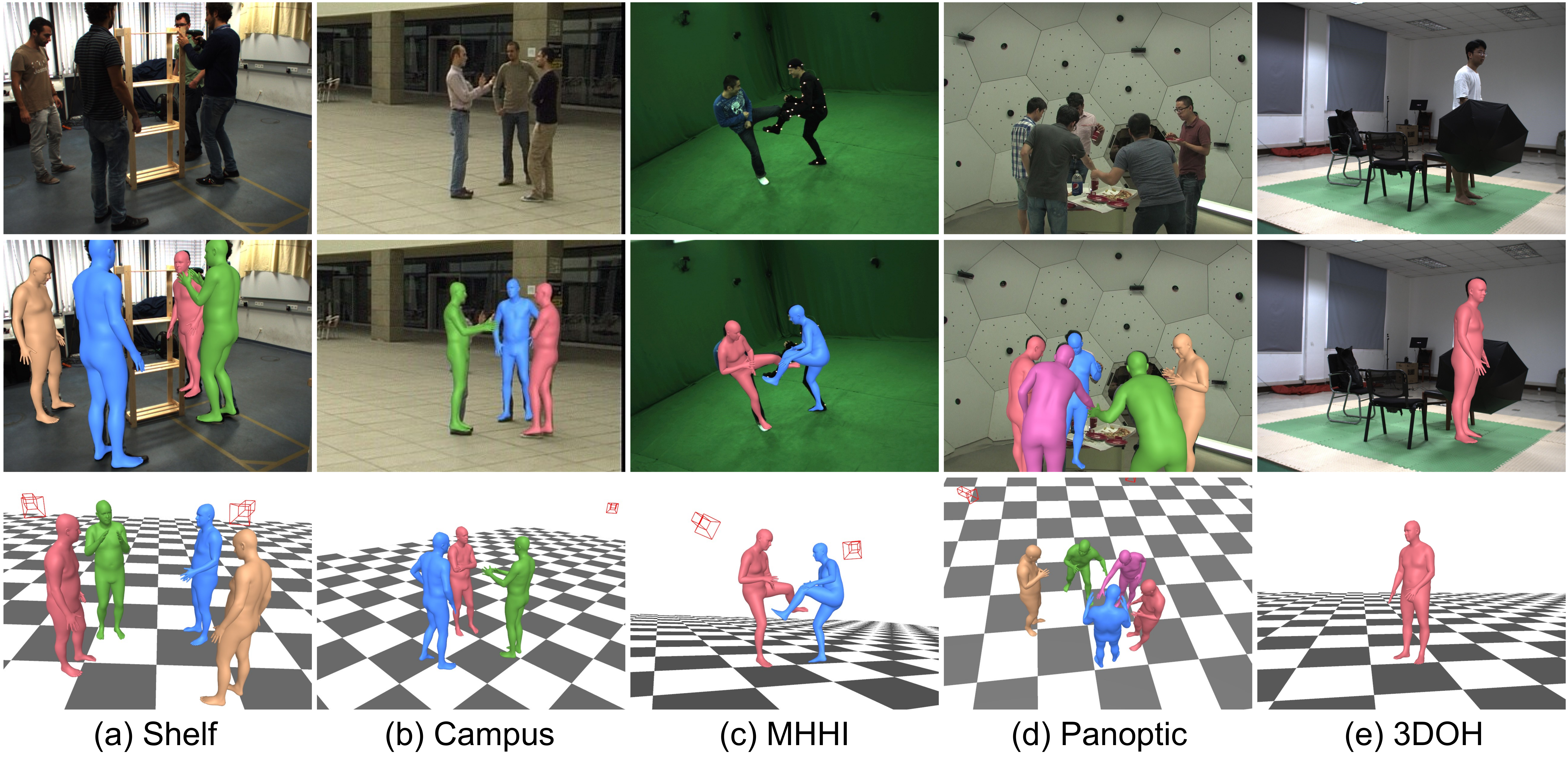}
    \end{center}
    \vspace{-7mm}
    \caption{The estimated results on different datasets. Our methods can obtain temporal coherent human motions and accurate extrinsic cameras parameters simultaneously from multi-view uncalibrated videos.}
    \label{fig:qulitative_results}
    \vspace{-7mm}
    \end{figure*}

\begin{table}
\begin{center}
    \resizebox{1.0\linewidth}{!}{
        \begin{tabular}{l|c c c|c c c}
        \noalign{\hrule height 1.5pt}
        \begin{tabular}[l]{l}\multirow{2}{*}{Method}\end{tabular}
        &\multicolumn{3}{c|}{\textit{Campus}} &\multicolumn{3}{c}{\textit{Shelf}} \\
        &A1   &A2  & A3 &A1 &A2  & A3 \\
        \noalign{\hrule height 1pt}
        Belagiannis~\etal~\cite{belagiannis20143d} &82.0 &72.4 &73.7 &66.1 &65.0 &83.2 \\
        Belagiannis~\etal~\cite{belagiannis20153d} &93.5 &75.7 &85.4 &75.3 &69.7 &87.6 \\
        Bridgeman~\etal~\cite{bridgeman2019multi}  &91.8 &92.7 &93.2 &99.7 &92.8 &97.7 \\
        Dong~\etal~\cite{dong2019fast} & 97.6 &93.3  &98.0 & 98.9 & 94.1&\textbf{97.8}  \\
        Chen~\etal~\cite{chen2020cross} &97.1 &\textbf{94.1} &98.6 &99.6 &93.2 &97.5 \\
        Zhang~\etal~\cite{zhang20204d} &-- &-- &-- &99.0 &96.2 &97.6 \\
        Chu~\etal~\cite{chu2021part} &\textbf{98.4} &93.8 &98.3 &99.1 &95.4 &97.6 \\
        VPoser-t~\cite{pavlakos2019expressive} &97.3 &93.5 &98.4 &\textbf{99.8} &94.1 &97.5 \\
        \textbf{Ours}&97.6 &93.7 &\textbf{98.7} &\textbf{99.8} &\textbf{96.5} &97.6 \\
        \noalign{\hrule height 1.5pt}
        \end{tabular}
    }
\vspace{-3mm}
\caption{Comparison with baseline methods that estimate multi-person 3D poses. The numbers are the percentage of correctly estimated parts~(PCP). The proposed method achieves state-of-the-art on some metrics. VPoser-t is a combination of VPoser~\cite{pavlakos2019expressive}.}
\label{tab:campus_shelf}
\end{center}
\vspace{-10mm}
\end{table}

\subsection{Multi-person Motion Capture}\label{sec:mocap experiment}
We first conducted qualitative and quantitative comparisons on Campus and Shelf datasets. To the best of our knowledge, no method has ever recovered human meshes on these datasets. We compared several baseline methods that regress 3D poses. \cite{belagiannis20143d} and \cite{belagiannis20153d} introduce 3D pictorial structure for multi-person 3D pose estimation from multi-view images and videos respectively. \cite{bridgeman2019multi, dong2019fast, chen2020cross, zhang20204d, chu2021part} are recent works based on calibrated cameras. The quantitative results shown in \tabref{tab:campus_shelf} demonstrate that our method achieves state-of-the-art on Campus and Shelf datasets in terms of PCP. Since only a few works target to multi-person mesh recovery task from multi-view input, we compared with EasyMocap~\footnote{https://github.com/zju3dv/EasyMocap} which fits SMPL model to the 3D pose estimated by \cite{dong2019fast}. Row 2 and row 4 of \figref{fig:qulitative_comparison} show that \cite{dong2019fast} produces the wrong result due to partial occlusion, while our method generates accurate poses with physics-geometry consistency. Besides, our method obtains more natural and temporal coherent results even for challenging poses since the proposed motion prior provides local kinematics and global dynamics.

We then evaluated our method on MHHI dataset. \cite{liu2013markerless,li2018shape,li2020full} can reconstruct closely interacting multi-person meshes from multi-view input, but all these works rely on accurate calibrated camera parameters. We conducted quantitative comparisons with these methods in \tabref{tab:MHHI}. The numbers are the mean distance with standard deviation between the tracked 38 markers and its paired 3D vertices in \textit{mm}. In the single-view case, since the motion prior provides additional prior knowledge, our method generates far more accurate results than \cite{li2018shape}. In addition, the proposed approach achieves competitive results with the least views.

To further demonstrate the effectiveness of the proposed method in single-view occluded situations, we show the qualitative results on 3DOH in \figref{fig:qulitative_results}. Our method can recover complete and reasonable human bodies from partial observation with the local kinematics and global dynamics in the motion prior. More qualitative and quantitative results on single-person datasets can be found in Sup. Mat.

\begin{table}
    
    \begin{center}
        \resizebox{1\linewidth}{!}{
            \begin{tabular}{l|c c c c c}
            \noalign{\hrule height 1.5pt}
            \begin{tabular}[l]{l}\multirow{1}{*}{Method}\end{tabular}
            
            &1 view  &2 views & 4 views&8 views &12 views  \\
            \noalign{\hrule height 1pt}

            Liu~\etal~\cite{liu2013markerless} & - &- &- & - & 51.67  \\
            Li~\etal~\cite{li2018shape} &1549.88&242.27&58.42 &48.57 &43.30  \\
            Li~\etal~\cite{li2020full} &- & 63.93&37.88 & 32.73&30.35  \\
            VPoser-t~\cite{pavlakos2019expressive} &158.33 &60.02 &38.46 &32.11 &31.48  \\
            \textbf{Ours}&\textbf{140.96} &\textbf{58.04} &\textbf{37.86} &\textbf{30.92} &\textbf{29.83}  \\

            \noalign{\hrule height 1.5pt}
            \end{tabular}
        }
    \vspace{-3mm}
    \caption{Quantitative comparison with multi-person mesh recovery methods on MHHI dataset. The numbers are the mean distance with standard deviation between markers and its paired 3D vertices in \textit{mm}.}
    \label{tab:MHHI}
    \end{center}
    \vspace{-6mm}
    \end{table}

\begin{table}
    \begin{center}
        \resizebox{1.0\linewidth}{!}{
            \begin{tabular}{l|c c c|c c c}
            \noalign{\hrule height 1.5pt}
            \begin{tabular}[l]{l}\multirow{2}{*}{Method}\end{tabular} &\multicolumn{3}{c|}{Panoptic Dataset} &\multicolumn{3}{c}{Shelf Dataset}\\
            
            &\textit{\begin{tabular}[1]{c}Pos.\end{tabular}} &\textit{Ang.} &\textit{Reproj.}&\textit{Pos.} &\textit{Ang.} &\textit{Reproj.}\\
            \noalign{\hrule height 1pt}
            PhotoScan         &505.02 &35.29 &188.18 &- &- &-   \\
            initial           &3358.51 &44.30 &637.21 &1532.42 &26.86 &79.34 \\
            w/o P-G consis. \textit{+ opt cam.}      &178.78 &1.10 &23.00  &29.09 &0.68 &18.88  \\
            VPoser-t~\cite{pavlakos2019expressive}  \textit{+ opt cam.}    &118.88 &0.64 &22.76 &34.30 &0.59 &18.83   \\
            MotionPrior \textit{+ opt cam.}            &\textbf{101.25} &\textbf{0.59} &\textbf{22.69}  &\textbf{23.18} &\textbf{0.52} &\textbf{18.70}  \\
            \noalign{\hrule height 1.5pt}
            \end{tabular}
        }
    \vspace{-3mm}
    \caption{Evaluation of the estimated camera. The \textit{Pos.} and \textit{Ang.} are position error and angle error between predicted cameras and ground-truth camera parameters. The units are \textit{mm} and \textit{deg}, respectively. The \textit{Reproj.} is re-projection error in pixel. The initial is the coarse camera parameters estimated from~\secref{sec:Preliminaries}. \textit{+ opt cam.} denotes simultaneously optimize cameras and human motions.}
    \label{tab:camera Calibration}
    \end{center}
    \vspace{-7mm}
    \end{table}

\begin{figure*}
    \begin{center}
    \includegraphics[width=1.0\linewidth]{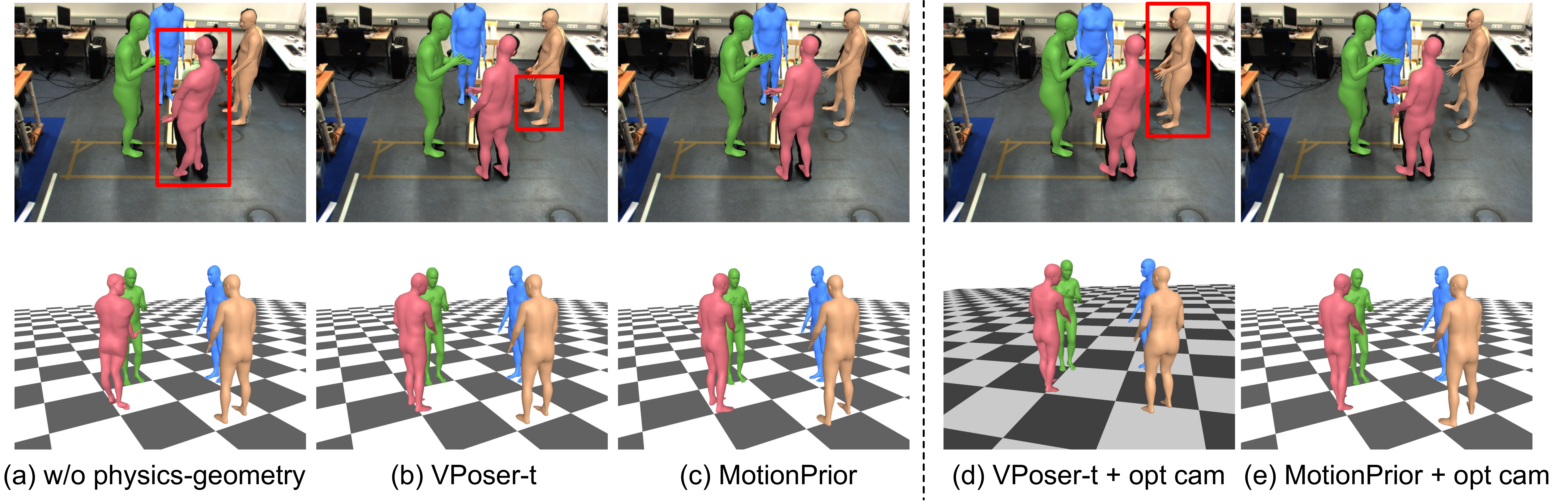}
    \end{center}
    \vspace{-7mm}
    \caption{Ablation on physics-geometry consistency and our motion prior. Without physics-geometry consistency, it can not obtain accurate motion due to the influence of noises. Since the lack of motion dynamics, the VPoser-t is hard to estimate plausible cameras and motions when the cameras are not provided. }
    \label{fig:ablation}
    \vspace{-5mm}
    \end{figure*}

\begin{table}
    \begin{center}
        \resizebox{1.0\linewidth}{!}{
            \begin{tabular}{l|c c|c c c}
            \noalign{\hrule height 1.5pt}
            \begin{tabular}[l]{l}\multirow{2}{*}{Method}\end{tabular}
            
            &\multicolumn{2}{c|}{\textit{MHHI}} &\multicolumn{3}{c}{\textit{Shelf}} \\
            &Mean   &Std  &A1 &A2  & A3 \\
            \noalign{\hrule height 1pt}
            VPoser-t~\cite{pavlakos2019expressive}  &31.48 &11.54 &99.8 &94.1 &97.5 \\
            w/o P-G consis. &32.31 &12.17 &92.4 &89.8 &91.6 \\
            w/o local linear &30.25 &11.07 &99.8 &95.4 &97.3 \\
            MotionPrior &\textbf{29.83} &\textbf{9.87} &\textbf{99.8} &\textbf{96.5} &\textbf{97.6} \\
            \hline\hline
            VPoser-t~\cite{pavlakos2019expressive} \textit{+ opt cam.} &43.72 &19.57 &97.4 &89.7 &89.7 \\
            w/o P-G consis.\textit{+ opt cam.} &49.34 &24.37 &91.5 &86.7 &88.6 \\
            w/o local linear \textit{+ opt cam.}&35.25 &17.07 &97.5 &90.4 &93.3 \\
            MotionPrior \textit{+ opt cam.}&\textbf{34.44} &\textbf{10.57} &\textbf{98.4} &\textbf{91.5} &\textbf{94.4} \\


            \noalign{\hrule height 1.5pt}
            \end{tabular}
        }
    \vspace{-3mm}
    \caption{Ablation on physics-geometry consistency and our motion prior. \textit{opt cam.} denotes simultaneously optimize cameras and human motions.}
    \label{tab:ablation}
    \end{center}
    \vspace{-10mm}
    \end{table}

\begin{figure}
    \begin{center}
    \includegraphics[width=1.0\linewidth]{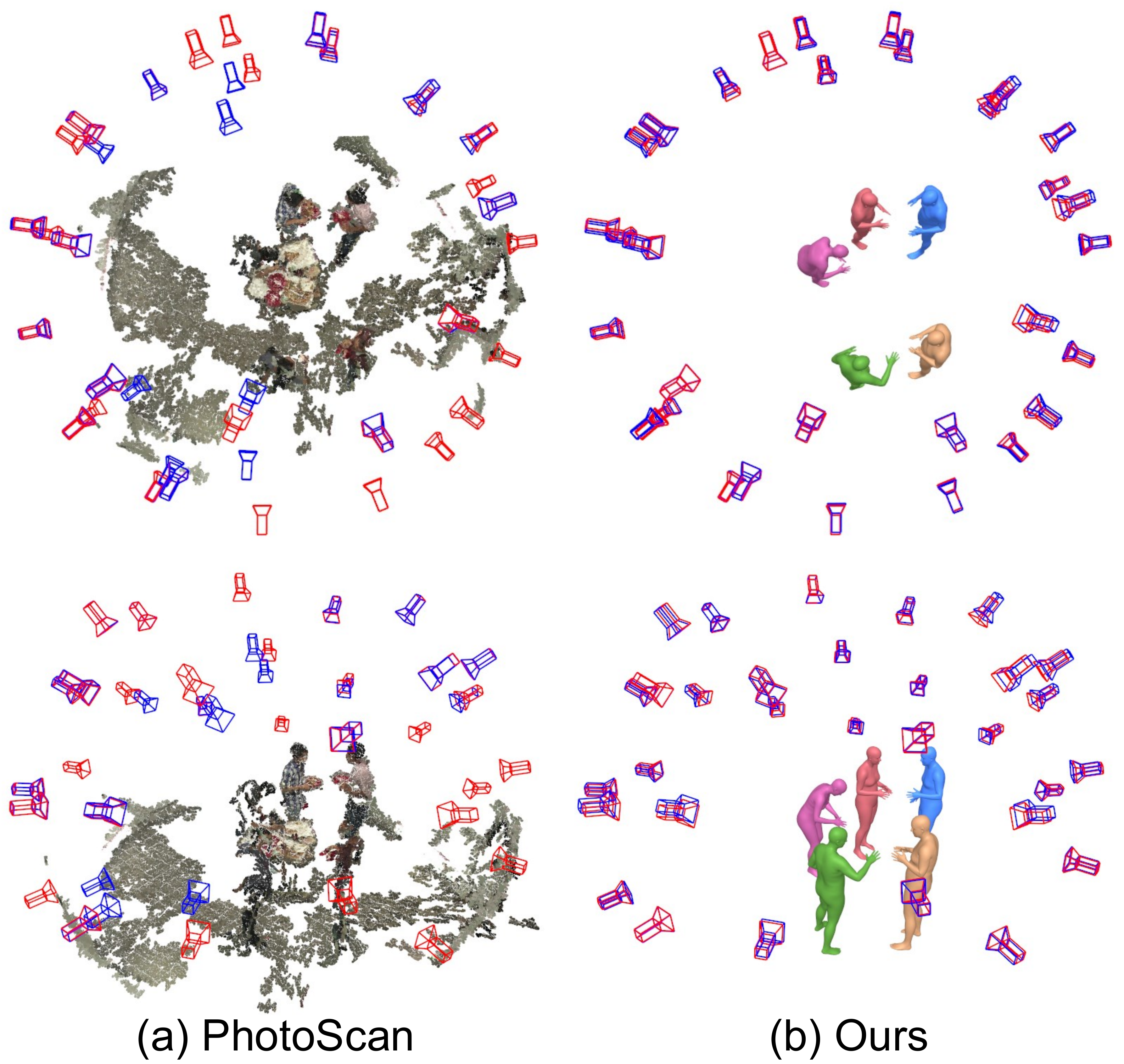}
    \end{center}
    \vspace{-6mm}
    \caption{PhotoScan can not work on sparse inputs. We conducted a comparison with PhotoScan on Panoptic with 31-views input. Our method accurately estimates all camera extrinsic parameters from noisy human semantics, while PhotoScan gets only a part of cameras.}
    \label{fig:qulitative_cam}
    \vspace{-6mm}
    \end{figure}

\subsection{Camera Calibration Evaluation}\label{sec:camera experiment}
We then qualitatively and quantitatively evaluate the estimated camera parameters. Since there exists a rigid transformation between the predicted camera parameters and the ground-truth provided in the datasets, we follow \cite{cioppa2021camera} to apply rigid alignment to the estimated cameras. We first compared with PhotoScan~\footnote{https://www.agisoft.com/}, which is a commercial software that reconstructs 3D point clouds and cameras. As shown in \tabref{tab:camera Calibration}, PhotoScan fails to work for sparse inputs~(Shelf dataset) since it relies on the dense correspondences between each view. We evaluate the results with position error, angle error, and re-projection error. Under relatively massive views, our method outperforms PhotoScan in all metrics. \figref{fig:qulitative_cam} shows the results on Panoptic dataset with 31 views. The cameras in red and blue colors are the ground-truth and the predictions, respectively. PhotoScan only captures part of the cameras with low accuracy. On the contrary, our method successfully estimates all the cameras with complete human meshes. We then compared with the initial extrinsic parameters estimated in \secref{sec:Preliminaries}. After joint optimization, the final results gain significant improvement. Our method achieves better performance both from massive and sparse inputs with the physics-geometry consistency and the motion prior.

\subsection{Ablation Study}\label{sec:ablation}
\noindent\textbf{Physics-geometry consistency}. We conducted ablation on the physics-geometry consistency to reveal its importance of removing the noises in the human semantics. \figref{fig:ablation} illustrates that without the consistency, the reconstruction is unnatural due to the noisy detections. As shown in \tabref{tab:ablation}, without the proposed consistency, the mean distance error of joint optimization increases 12.42, demonstrating its significance.

\noindent\textbf{Motion prior}. VPoser-t is a combination of \cite{pavlakos2019expressive} which lacks global dynamics. We first compared it to illustrate the superiority of the proposed motion prior. \tabref{tab:ablation} shows that the standard variance of our method on MHHI is smaller since the motion prior models the temporal information. \tabref{tab:camera Calibration}, \tabref{tab:ablation} and \figref{fig:ablation} demonstrate that due to the lack of temporal constraints, VPoser-t is more sensitive to the noisy detections. The local linear constraint ensures a smooth transition between each frame of the latent code. We then removed the local linear constraint when training the motion prior. In \tabref{tab:ablation}, without local linear constraint, although the mean distance error of joint optimization on MHHI dataset is small, the standard variance of which is large. Thus, the results prove that the constraint is effective in modeling temporal coherent motions.

%% file: conclusion.tex
\section{Conclusion}\label{sec:conclusion}
This paper proposes a framework that directly recovers human motions and extrinsic camera parameters from sparse multi-view video cameras. Unlike previous work, which fails to establish view-view and model-view corresponds, we introduce a physics-geometry consistency to reduce the low and high frequency noises of the detected human semantics. In addition, we also propose a novel latent motion prior to jointly optimize camera parameters and coherent human motions from slightly noisy inputs. The proposed method simplifies the conventional multi-person mesh recovery by incorporating the calibration and reconstruction into a one-stage optimization framework.

\noindent\textbf{Acknowledgments.}
The authors would like to thank Professor Yebin Liu and Professor Kun Li for sharing the data. This work was supported in part by National Key R$\&$D Program of China (No. 2018YFB1403900), in part by National Natural Science Foundation of China (No. 61806054), in part by Natural Science Foundation of Jiangsu Province (No. BK20180355), Young Elite Scientist Sponsorship Program by the China Association for Science and Technology and "Zhishan Young Scholar" Program of Southeast University.